\documentclass[12pt,a4paper]{article}
\usepackage[hyperref]{acl2020}
\usepackage{times}
\usepackage{subfigure}
\usepackage{latexsym}
\usepackage{dirtytalk}
\usepackage{dblfloatfix}
\usepackage{graphicx}
\usepackage{amsmath}

\usepackage{microtype}

\aclfinalcopy 

\title{Adversarial Examples Generation for Reducing Implicit Gender Bias \\in Pre-trained Models}

\author{
  Wenqian Ye \\
  \texttt{wy2029@nyu.edu}
  \\\And Fei Xu\\
  \texttt{jx1260@nyu.edu}
  \\\AND Yaojia Huang\\
  \texttt{yh3276@nyu.edu}
  \\\And Cassie Huang\\
  \texttt{sh4125@nyu.edu}
  \\\And Ji A\\
  \texttt{ja3802@nyu.edu}
}

\date{}

\begin{document}
\maketitle
\begin{abstract}

\noindent
Over the last few years, Contextualized Pre-trained Neural Language Models, such as BERT, GPT, have shown significant gains in various NLP tasks. To enhance the robustness of existing pre-trained models, one way is adversarial examples generation and evaluation for conducting data augmentation or adversarial learning. In the meanwhile, gender bias embedded in the models seems to be a serious problem in practical applications. Many researches have covered the gender bias produced by word-level information(e.g. gender-stereotypical occupations), while few researchers have investigated the sentence-level cases and implicit cases. 

In this paper, we proposed a method to automatically generate implicit gender bias samples at sentence-level and a metric to measure gender bias. Samples generated by our method will be evaluated in terms of accuracy. The metric will be used to guide the generation of examples from Pre-trained models. Therefore, those examples could be used to impose attacks on Pre-trained Models. Finally, we discussed the evaluation efficacy of our generated examples on reducing gender bias for future research.

\end{abstract}

\section{Introduction}

The attention paid on the field of machine learning systems has kept increasing in recent years. However, many machine learning models are shown to be vulnerable to adversarial attacks; namely, these models would give incorrect predictions by feeding them intentionally-modified inputs (aka adversarial examples). 
For example, \citet{belinkov2018synthetic} showed that swapped characters and typos lower the BLEU score significantly in neural machine translation models. On the other side, the adversarial attacks generated on the model developers' side will make the trained models more robust.  

Most recently, with the increasing focus on large-scale deep learning systems, a plethora of pre-trained models have emerged extensively and shown significant gains in various NLP tasks with fine-tuning. Such remarkable progress intrigued researches on evaluating whether these models are robust by conducting adversarial attacks and, in the meantime, provided suggestions (e.g. data augmentation, adversarial learning) to improve the performance of existing pre-trained models. In practice, the propagation of gender bias in pre-trained models has raised 
the public awareness, and it may pose the danger of reinforcing damaging stereotypes in downstream applications. Models automating resume screening have also been proved to have a heavy gender bias favoring male candidates (\citet{lambrecht}). Such data and algorithmic biases have become a growing concern.

The state-of-the-art techniques, like the one proposed by \citet{bolukbasi2016man}, have already covered word-level gender bias detection using the gender subspace distance. Also, from the result of \citet{bordia-bowman-2019-identifying}, it has shown the possibility to accurately measure the word-level gender bias in a text corpus and reduce such bias by proposing a regularization loss term for
the language model. However, these word-level techniques cannot generalize well to phrase-level or sentence-level cases. For existing techniques on the sentence-level bias like \citet{may-etal-2019-measuring}, they also relied on the word-level techniques and have a strict limitation on the sentences. In this project, we proposed a generalized method to generate implicit gender bias samples in sentence level and we proposed a metric to measure such bias. The metric can guide the generation of bias examples from Pre-trained models. And the creation of such examples allows for comparison of its impact on different Pre-trained models. Finally, we discussed our plan to evaluate the efficacy of our generated examples on reducing gender bias, we left this to our future investigation due to the limited time.

\section{Methods}

\subsection{Problem Statement}\label{2.1}
Most previous work on gender bias in the NLP community has focused on identifying the bias and debiasing existed models. To tackle the problem at its root and allow for a comparison of its impact on various language models, we believed that an open dataset must be created.

To that end, the first step is to quantify the qualitative definition of implicit gender bias. Inspired by the gender bias taxonomy proposed by \citet{hitti2019proposed}, we focused on two main gender bias issues: structural bias and contextual bias, because they have not been well studied and needed contextual information to understand it. In general, we defined implicit gender bias in the text as the whole sentence that connotes, prejudices, or implies an inclination against one gender. Implicit gender bias cannot be indicated only by the words used in the sentence. We also introduced a metric to quantify it. The details are shown below.

\textbf{Structural bias.} Followed by the definition of  \citet{hitti2019proposed} as well as the examples in \citet{macaulay1997don}, we defined structural bias as one type of implicit gender bias. This kind of gender bias occurs when bias could be traced down from a particular syntactical construction. 
This includes searching for any syntactic patterns that enforce gender assumptions in a gender-neutral context. In another word, this type of gender bias appears when a gendered pronoun is linked to a gender-neutral term in a gender-free context. Gendered pronouns included:\emph{ he, his, him, himself, she, her, hers and
herself}.
\begin{itemize}
    \item \say{Someone winds up his right arm and knocks the fighter down with a haymaker.} - shows an action of a person and assumes the person to be a man by referring to \say{his}.
    \item \say{The belly dancer dances on stage shaking her hips and body.} - gives a fact about an arbitrary dancer and assumes a woman to be the dancer by referring to \say{her}.\\\\
    \emph{Counter example:}
    \item \say{A girl will always want to play with her Barbie doll.}- although describing a stereotype,
it is not assuming the gender for a gender
neutral word because the word girl (gendered -
female) is linked to a female pronoun. Therefore, it is
not gender bias.
\end{itemize}

\textbf{Contextual bias.}
On the other side, contextual gender bias does not have a pattern-based definition. It often requires the learning of the link between gender obvious keywords and contextual information. Unlike above type, this type of bias cannot be
observed through syntactical structure but requires contextual knowledge and some degree of human perception.
\begin{itemize}
    \item \say{Presidential candidates need their wives to support them throughout their campaign.} - the word
\say{wife} is described as a supporting role when
we do not know the gender of the Presidential candidates and the supporting role can be a male partner, a
husband.
    \item \say{Your sister could go to College, but would she get a degree?} - it clearly casts doubt on
 the possibility of academic success for this woman.
    \item \say{John must love football because all boys like
playing with it.} - assumes that football are
only liked by boys.
\end{itemize}
\textbf{Implicit gender bias metric.} \citet{Caliskan_2017} first introduced Word Embedding Association Test (WEAT) to quantify gender bias in English embedding, which is a permutation test. The test considered two sets of target words and two sets of attribute words, and measured the association between them. Formally, let X and Y be two sets of target word vectors of equal size (such as word embeddings related to different professions) and let A and B be sets of
attribute word vectors (e.g., gender marked definition word vectors,i.e, \say{man}, \say{woman}). Let $cos(\vec{a},\vec{b})$
denotes the cosine similarity between word vectors $\vec{a}$ and
$\vec{b}$. The test objective function is a difference between sums over the each target words,
$$ s(X,Y,A,B) = \sum_{x \in X}s(x, A, B) -\sum_{y \in Y}s(y,A,B)$$
where $s(\vec{w}, A, B)$ measures the difference of the word $w$ with the attributes, 
\begin{align*} 
    & s(\vec{w}, A, B)=  \\ & \frac{1}{|A|}\sum_{\vec{a} \in A}cos(\vec{w},\vec{a}) -\frac{1}{|B|}\sum_{\vec{b} \in B}cos(\vec{w},\vec{b}).
\end{align*} 

Followed by this work, we extend the target word sets by adding a kind of phrase sets. We considered gender-neutral behavioral phrases but usually depicted in stereotypes(like \say{knocks the fighter down} and \say{wear a revealing dress}). This metric will be used to compute the bias score of our produced examples by our filtering method and we discussed using this as a guidance to generate gender bias samples from Pre-trained models in the future work.

\begin{figure*}[h]
\centering
  \centering
  \includegraphics[width=1\linewidth]{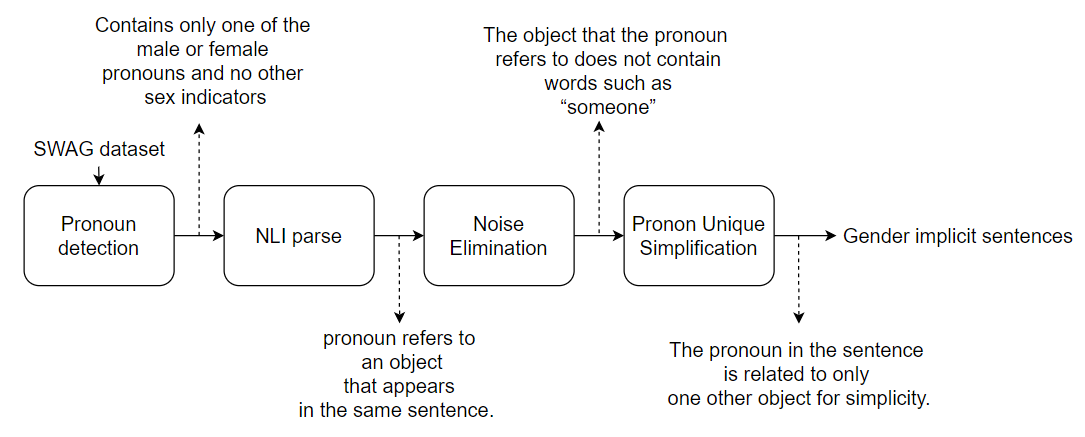}
  \caption{\textbf{Design diagram for filtering the sentences with implicit gender bias.} }
   \label{filter1}
\end{figure*}

\subsection{Approach}
Between the two types of gender bias proposed in section 2.1, we mainly focused on judging Structural gender bias. Our method of examining whether a given pre-trained model exhibit implicit gender bias by feeding the model sentences where gendered pronouns are masked and comparing the probabilities it predicts for male words and female words. We designed a two-stage method that contains a filter stage and a bias computation stage. In the filter stage, we perform a 4-step filtering to get desired sentences. We propose such algorithm to ensure a sentence contains only one gendered pronoun and no other sex indicators so that the pre-trained model we try to investigate will not be misled by irrelevant words in the sentence.

\section{Experiments}
\subsection{Initial Datset}
We chose SWAG, a large-scale adversarial dataset for grounded commonsense inference \cite{zellers2018swagaf}, to be our starting point. Each row of SWAG dataset has the format as shown in Table \ref{SWAG}, where the fields in the first column are the column headers in SWAG.

\begin{table}[h]
\begin{tabular}{|l|l|l|l|l|}
\hline
\textbf{video-id} & lsmdc1005\_Signs-5092\\
\hline
\textbf{fold-ind}  &   6072   \\
\hline
 \textbf{startphrase}     &    Someone walks over to the\\ &radio. Someone\\
 \hline
 \textbf{sent1}& Someone walks over to the \\&radio.\\
 \hline
 \textbf{sent2}&Someone\\
 \hline
 \textbf{gold-source}&gold\\
\hline
...&...\\
\hline
\end{tabular}
\caption{\textbf{The Format of The Training Data of SWAG.} The rows represents the original columns in SWAG and only the first 6 columns are shown here.}
\label{SWAG}
\end{table}

We extracted \textbf{sent1} from the dataset because we only need one complete sentence for each sample. We eventually obtained $73547$ lines of sentences.

\subsection{Filter Stage}
The dataset we got includes various types of sentence, but not all of the sentences exhibit gender bias. For example, the sentence ``someone walks over to the radio'' in Table \ref{SWAG} does not contain any words related to gender. Therefore, we proposed a filtering process to get sentences that could be modified to exam gender bias in pre-trained models. The filtering steps are shown in Figure \ref{filter1}.

We defined a group of gendered identities such as ``sister'', ``boy'', ``actress'', etc. The words in gendered pronouns and gendered identities are collectively called sex indicators. Then, we selected sentences that contain only one of the gendered pronouns and no other sex indicators. By doing so, we eliminated sentences with no gendered pronoun and sentences containing multiple sex indicators such as ``the boy met his friends in his house.'' The latter is not wanted because a model can infer the gender of one word from the other sex indicators.

We used the model proposed by \citet{Lee2017EndtoendNC}, which achieves an F1 score of $78.87$ on the Ontonotes $5.0$ dataset, to do coreference resolution on the sentences. We obtained one or more groups of associated expressions for each sentence. For example, the correct result for ``the nurse is looking after her patients'' is a group containing ``the nurse'' and ``her''. For each sentence, we checked whether there exists a bag that satisfies the following criteria:

\begin{itemize}
    \setlength\itemsep{0em}
    \item the number of expressions in the bag is $2$,
    \item one expression is a gendered pronoun and the other is a word other than ``someone''.
\end{itemize}

These criteria are not mandatory as they only serve to formulate our dataset in a more restrictive fashion. The sentences not fulfilling the above criteria were filtered out from the dataset.

Finally, we masked the gendered pronouns in the sentences.

\subsection{Bias Computation Stage}
\begin{figure}[htbp]
\centering
  \centering
  \includegraphics[width=0.8\linewidth]{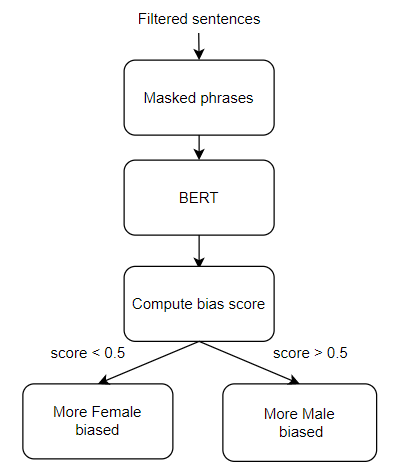}
  \caption{\textbf{Design diagram for computing bias} }
   \label{filter}
\end{figure}

After acquiring the masked sentences that cannot be explicitly inferred by word level techniques, we used them to test how gender biased the Pre-trained model is. Figure \ref{filter} shows the procedure to do such job. We use Pre-tranied model to do a masked model prediction on those sentences to get top $k=10$ predictions and their corresponding probabilities. From those predictions, we picked the highest probability for the word that refer to a female gender identity as $P(s|f)$. In the same way, we picked the highest probability for the word that refer to a male gender identity as $P(s|m)$. We define the function to compute bias score $bias(s)$ as:

\begin{equation}
    \text{bias}(s) = \frac{P(s|m)}{P(s|m) + P(s|f)},
\end{equation}
where $s$ is the masked sentence.

An online Masked Language Modeling tool provided by AllenNLP\cite{allen} is used here to show an short example. The model used here is BERT, the first large transformer to be trained on this task. The following masked sentence is used as an input for this mode:

\begin{itemize}
\item \say{[MASK] stands up next to someone.}
\end{itemize}

The model predicts [MASK] to be 'He' with a probability of $43.5\%$, and 'She' with a probability of $19.5\%$. According to our metric for computing bias, the score of the prediction for this sentence is 0.69, which is biased to male on this scenario.

We fed a masked sentence to the model and calculated the score for the prediction. If the score does not fell in the range $[0.5-\delta, 0.5+\delta]$, in which $\delta$ is the threshold we defined, we considered the model is gender\-biased on the input sentence.

\section{Results and Analysis}
By filtering the SWAG dataset, we got $663$ sentences that could be used to examine possible implicit gender bias in pre-trained model. We fed those sentences into BERT-uncased model, BERT-cased model and DistilBERT to do further validation. 

\subsection{Evaluation and Case study}
To prove our method to filter sentences with structural gender bias is reasonable, we first examined the filtered sentences by ourselves in terms of accuracy,
$$Acc = \frac{\text{\# of gender bias samples}}{\text{\# of all filtered samples}}.$$

\begin{table}[h]
\begin{tabular}{|l|l|}
\hline
\textbf{Statistic} & \textbf{Numbers}\\
\hline
No. of original examples & 73547\\
\hline
No. of filtered examples  &   663   \\
\hline
No. of gender-bias examples   &    602\\
\hline
Accuracy     &    90.79\%\\
\hline
\end{tabular}
\caption{Accuracy of our filtered method to produce gender-bias samples.}
\label{acc result}
\end{table}
We sampled $663$ examples from the original dataset, and we labeled $602$ out of $663$ filtered results as gender-bias samples. The filtered results were reviewed by $4$ of our group members and we took agreements at least $3$ members as truly gender-biased for reliability. The details was shown in Table \ref{acc result}.

We further gave some case studies of our results. 
We found that the
non-structural bias sentence, like the counter example from Section $2.1$: \say{A girl will always want to play with her Barbie doll.} was successfully filtered out by our method.
The two correct examples with structural bias from Section $2.1$ was picked from the result we got after the filtering stage. We masked one of the sentence like the following: 

\begin{itemize}
    \item \say{Someone winds up [MASK] right arm and knocks the fighter down with a haymaker.}
\end{itemize}

The masked sentence was then fed into two pre-trained models. The BERT-uncased model predicted [MASK] to be 'his' with a probability of $14.2\%$, and 'her' with a probability of $11.3\%$. The DistilBERT model predicted [MASK] to be 'his' with a probability of $54.2\%$, but 'her' with only probability of $3.05\%$. According to our metric for computing bias, BERT-uncased model get a score of $0.56$ on this sentence, while DistilBERT model got a score of $0.95$. This result indicates that for a sentence produced by our method, the two pre-trained models showed different level of implicit gender bias, with DistilBERT model having a more severe bias.

\subsection{Other Analysis}
\begin{figure*}[t!]
  \centering
  \begin{subfigure}[\textbf{Distribution of the result from BERT-uncased}]{
    \includegraphics[width=0.48\textwidth]{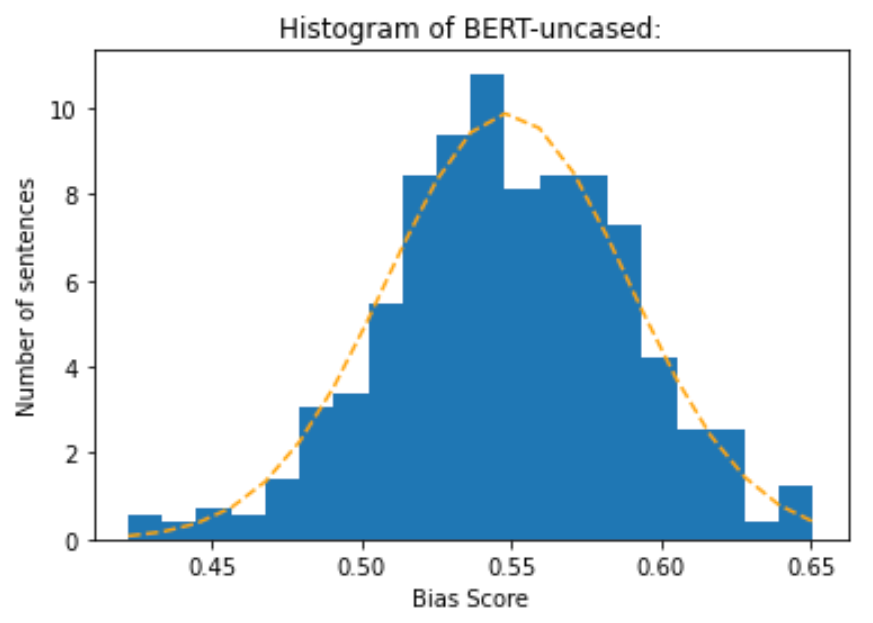}
    \label{uncased}}
  \end{subfigure}
  \hfill
  \begin{subfigure}[\textbf{Distribution of the result from BERT-cased}]{
    \includegraphics[width=0.48\textwidth]{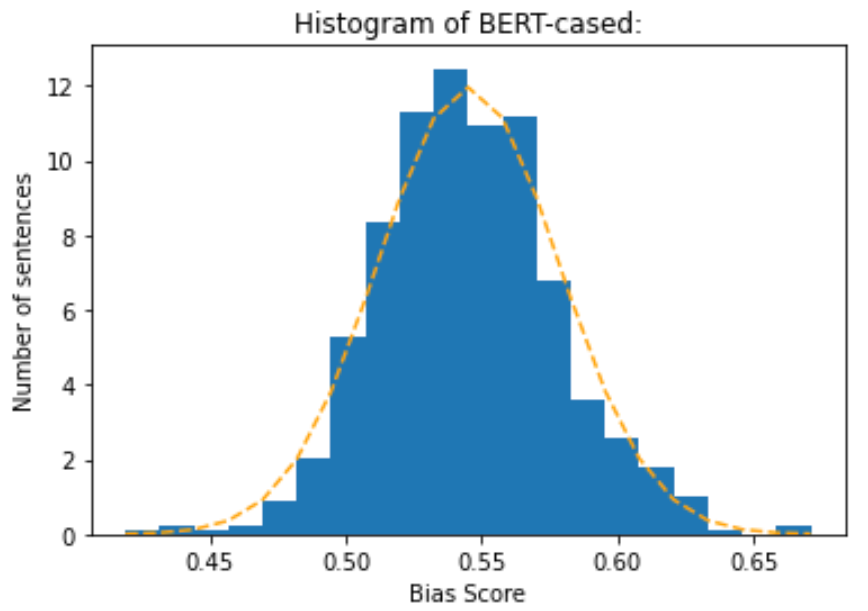}
    \label{cased}}
  \end{subfigure}
 \caption{\textbf{Distributions of the results obtained from two models}. Both of the distributions have a Gaussian shape and have peaks around $0.55$, indicating the model is more biased to male on our dataset.}
 \label{both}
\end{figure*}
The score distributions of BERT-uncased and BERT-case are shown in Figure \ref{both}.

In BERT-uncased, the result shows that, among those sentences, $542$ of them are male biased, with an average bias score of $0.556$. The other $65$ sentences are female biased, with an average bias score of $0.47$. The rest $33$ sentences are either giving a gender identity possibility less than $5\%$, or just failed to predict any word with gender identity by BERT-uncased. 

In BERT-cased, among $663$ filtered sentences, the result shows that among those sentences, $575$ of them are male biased, with an average bias score of $0.550$. The other $43$ sentences are female biased, with an average bias score of $0.483$. The rest $45$ sentences are either giving a gender identity possibility less than $5\%$, or just failed to predict any word with gender identity by BERT-cased. 

According to Figure \ref{both} and our rule to judge if a model has gender bias, we can conclude that BERT is considered to be more likely to predict a masked token to be a word with a male identity on our dataset.

\section{Related Works}
\textbf{Robustness of pre-trained models.} A lot of works have been proposed to study the robustness of pre-trained models in recent years. 
One spectrum of works approaches the attack in a black-box manner, focusing on variation of adversarial example generation (\citet{Robust?,kaushik2019learning}). 
Another spectrum of works attacks the model in a white-box way which uses the model information to generate adversarial examples\cite{liu2020adversarial,tu2020empirical}. Our work mainly focused on evaluate the robustness of Pre-trained models on a particular aspect:gender-bias. 

\textbf{Gender bias in Language models.} 
Neural language models are becoming prevalent in real world applications, and embedded bias will have a great societal impact\cite{buolamwini2018gender, tonry2010social}. Many types of research have studied the gender bias in neural language models such as word embedding models \cite{bolukbasi2016man, Caliskan_2017, wang2020double} and NMT models\cite{saunders2020reducing}. \citet{zhao2020gender} investigated the Gender Bias in Multilingual Embeddings and Cross-Lingual Transfer. \citet{may2019measuring} measured gender bias in sentence encoders such as ELMo and BERT. \citet{zhao2017men} showed the model can amplify the bias without appropriate adjustments. In \citet{shah2019predictive} 's work, they summarized the previous works and proposed a unifying predictive bias framework for NLP. Different from the previous work, our work aim at understanding the implicit bias in phrase-level or sentence-level and proposed a model to generate bias examples to attack the pre-trained models.

\section{Conclusion and Future work}
In this paper, we investigated implicit gender-bias text at sentence-level. We defined the implicit gender-bias as the whole sentence that connotes, prejudices against one gender. In particular, we focused on two main gender-bias issues which have not been well studied and needed contextual information to identify it. In the meantime, a metric was introduced to compute the gender-bias score of a sentence. The purpose of our project is to create an open dataset allows for a comparison of its impact on different language models and to reduce gender-bias if they existed in those models. To the end, we presented a two-stages method to produce such implicit samples from an dataset. Finally, we evaluated the results given by our method. All filtered results were reviewed by $4$ members. A sample which labeled as gender-bias only if at least $3$ annotators agree on the result. A detailed analysis of the results was also presented to see the two types of gender-bias studies by us. We left the gender-bias score computed by the proposed metric in our future study, to see if it is correlated with the results computed by our method. There are also four other directions need to further investigate,

\begin{itemize}

\item We should test our approach on more dataset to get a more comprehensive validation of results. Because our method to test gender bias so far on a pre-trained model depends on the dataset we used to filter sentences.

\item We should test our approach on more pre-trained models like the variants of transformers, ALBERT or RoBERTa.

\item With the help of proposed metric, automatic generating more implicit gender-bias examples using various contextualized Pre-trained models. 
\item Last but not least, we should evaluate the efficacy of our generated examples on reducing gender bias of different contextualized language models.
\end{itemize}

\section{Acknowledgements}
We are grateful to Prof. He He for reviewing our proposal and giving us helpful advice on this project. 

\bibliography{anthology,acl2020}
\bibliographystyle{acl_natbib}

\end{document}